 \documentstyle[12pt]{article}
\title{Reconstruction of~3-D Rigid Smooth Curves Moving Free when Two  
Traceable Points Only are Available }              

\author{Mieczys{\l}aw A. K{\l}opotek\\
        \em Institute of Computer Science Polish Academy of Sciences }

\date{}

\newcommand{\Bem}[1]{}
\begin{document}
\setlength{\unitlength}{1cm}

\maketitle

\section{Introduction}
     Reconstruction of shape of surrounding  objects  is  a  vital 
task to cope with by future (and partially by present) intelligent 
systems interacting with real world.  In  general,  recovering  of 
shapes  of  3D  space   objects   from   2D   images  
\cite{B:1,B:2,B:3,B:4,K:8,M:16,M:17,R:20,R:21,Y:28} 
has not been very successful as this task  is  too 
under-constrained (unless shape  \cite{H:5,S:23}, shading etc  
\cite{K:7,S:22} clues 
are available). Hence practical applications are rather  based  on 
sensing (via laser beams, ultrasonic methods etc. \cite{M:18,M:19}).  Though 
successful in recovering shapes  of  surfaces,  sensing  fails  to 
reconstruct curve-shaped objects as well as curved surface edges.

     So this remains still a competition area  for  2D  projection 
based recognition methods. Some promising  results  were  in  fact 
achieved in recovering objects from multiframes (a  time  sequence 
of projections of the moving object)  \cite{J:6,L:13,U:24,W:25,W:26} 
as  this  task 
is over-constrained. Also in  cases  where  features  of  interest 
cannot be all traced from  frame  to  frame ---  e.g.  smooth-curve 
shaped objects   \cite{K:9,K:10,K:11,L:12,L:14,L:15,W:27}. 
 In  fact,  only  several 
points (usually end points) are traceable, and the remaining  ones 
are  not.  The  strategy   consists   usually   of   two   stages: 
reconstruction of space parameters of traceable points, thereafter 
reconstruction of non-traceable points. 

     This paper extends (in sections 3 \& 4)  previous  results  in 
that  sense  that  for  orthogonal  projections  of  rigid  smooth 
(true-3D) curves moving totally free  it  reduces  the  number  of 
required traceable points to two only (the best results  known  so 
far to the author are 3 points from free motion and 2 for  motion 
restricted to rotation around a fixed  direction  and  and  2  for 
motion restricted to influence of a homogeneous force field).  The 
method  used  is  exploitation  of   information   on   tangential 
projections.  Section  5  contains  a  remark  on  possibility  of 
simplification of reconstruction of flat curves  moving  free  for 
prospective projections. 

\section{Previous Work}

     The following table summarizes previous work in the  area  of 
reconstruction of rigid  curves  from  multiframes  under  various 
shape and  motion  restrictions  for  orthogonal  and  prospective 
projections. Shift/rotation motion is ``uniform''  if  in  the  same 
elapsed time the same amount of shift/rotation occurs. The  motion 
is free if it does not  fit  this  requirement  of  uniformity.  A 
homogeneous force field causes the mass center  point  to  have  a 
constant acceleration vector. 

\begin{tabbing}
{\it free motion, bounded by homogeneous  }\=    {\bf traceable points \quad} 
     \= {\bf   of Frames \quad  }\=     rence     \kill
{\bf Motion Type }\> {\bf    Number of } \> {\bf     Number } \>  {\it    
Refe- }\\          \>  {\bf traceable points }  \> {\bf of Frames  }\>   {\it 
rence }\\\\
{\it FLAT (2D)  CURVES IN 3D }\\
{Orthogonal Projection }\\
{\it free motion}   \>  {\it 2}  \> {\it 2}  \>    [12] \\
{Prospective Projection }\\
{\it free motion}  \>  {\it 3} \>   {\it 3 }  \>  [11] \\
\\
REAL 3 D CURVES \\
Orthogonal Projection: \\
{\it uniform rotational motion}  \>  {\it 2} \>  {\it 4} \>   [27, 12] \\
{\it free rotation around a }\\
{\it \quad   fixed direction } \>  {\it 2}  \> {\it  4} \>     [9]\\
{\it free motion}  \> {\it   3} \> {\it  3 } \>     [9, 10]\\
{\it free motion, bounded by homogeneous  }\\
{\it  \quad   force field }\>  {\it 2} \> {\it  n } \>      [10]\\
Prospective  Projection:  \\
{\it rotation-free motion } \>  {\it 2 }\> {\it   2 }\>   [9] \\
{\it uniform rotational motion } \> {\it 2 } \>{\it    5 }\>   [9] \\
{\it free rotation around a }\\
{\it  \quad   fixed direction}  \> {\it  3} \> {\it   3 } \>  [9] \\
{\it free motion  }  \> {\it   4}  \> {\it   3 }\>    [9] \\
{\it free motion, bounded by homogeneous }\\
{\it \quad   force field } \> {\it  2} \>{\it   n } \>  [11] \\
Stereoscopic   Vision:\\
{\it free motion }\>  {\it  2 }\> {\it 1 }\>  [9, 28] 
\end{tabbing}

\input{SZENE1}
\section{Reconstruction of Traceable Features}
     Let us characterize the traceables of the  smooth  curve.  We 
assume that we can trace two points (usually endpoints) of it. Let 
the traceable points be $A$ and $B$. Their projections  be  called  
$A'_{i}$ 
and $B'_{i}$ ($i$ -- frame index).  Both $A'_{i}$ 
and $B'_{i}$ are observables. 
Furthermore we  can  observe  the  angles  between  $A'_{i} B'_{i}$  and  the 
projections of tangentials at $A$ and $B$ being projections of  angles 
between $AB$ and tangentials themselves (Fig.1).
 Let us  call  $\alpha$  the  angle 
between $AB$ and the tangential at $B$, and $\beta$ the angle between $AB$ 
and tangential at $A$. $\phi$  be  the  angle 
between the plane containing $AB$ and tangential at $A$ and the  plane 
containing $AB$ and the tangential at $B$.
 
     Length of $AB$ be called c. c,  $\alpha$, $\beta$, $\phi$    are  fixed  
through  all frames. 

     Let us consider the relation between the $i^{th}$    frame  and  the 
curve -- especially the line $AB$ and the tangential  at  $B$.  We  can 
always imagine that the current position of the curve was achieved 
as follows:

\input{SZENE2}

\input{SZENE3}

\input{SZENE4}

\begin{enumerate}
\item 
At the beginning $A$,$B$ and tangential at $B$ lay in the frame plane 
in such a way that $A_{i}' \ = \ A$. Let us draw a straight line $l1$  
through in the frame place perpendicular  to  $AB$.  Let  $p1$  be  the  
plane perpendicular to the frame plane and containing the line $l1$. 
(Fig.2).
\vspace*{-0.3cm} 
\item
First the curve is rotated by an angle $\delta_{i}$   around  the  by  now 
line $AB$ (Fig. 3.).
 Let us fix on the tangential at $B$ the point $S$ at which by  now 
the tangential crosses the plane $p1$.  Let  $S'$  be  the  orthogonal 
projection of $S$  in the plane $p1$ onto the line $l1$. Then  we  have: 
$\angle AS'S \ = \ 90 ^{o} $, $\angle SAS' \ = \ \delta_{i}$ ,
$\angle ABS \ = \ \alpha$,  hence:
\vspace*{-0.3cm}
\begin{itemize}
\item[1)]
$ \frac{AS}{AB} \  = \ \mbox{tg}  \alpha$
\item[2)]
$ \frac{AS'}{AS} \ = \ \cos \delta_{i}$
\item[3)]
$ \frac{SS'}{AB} \ = \ \sin \delta_{i}$
\end{itemize}
\vspace*{-0.3cm}
\item
Thereafter we rotate the whole curve together with the point  $S$ 
(not $S'$) around the line $l1$ by the angle $\tau_{i}$ (Fig.4.). 
Let $S''$ be  orthogonal 
projection of the newly positioned $S$ onto the  frame  plane.  Then 
obviously $\angle BAB' \ = \ \tau_{i}$ , 
$\angle S''S''A \ = \ 90^{o}$ , $\angle SS'S'' \ = \ 90 ^{o}  - \tau_{i}$ . 
Hence:\vspace*{-0.3cm}
\begin{itemize}
\item[4)]
$ \frac{AB'}{AB} \  = \ \cos \tau_{i}$
\item[5)]
$ \frac{S'S''}{SS'} \  = \ \cos (90^{o}  - \tau_{i})$
\end{itemize}
\vspace*{-0.3cm}

\noindent
Let us denote by $D'$ the crossing point of the lines $l1$  and  $B'S''$. 
As we know the line $l1$  and  the  direction  of  $B'S''$  (being  the 
orthogonal projection of the tangential $BS$ at $B$), we know also the 
position of $D'$. We obtain:
\vspace*{-0.3cm}
\begin{itemize}
\item[6)]
$  AS' \ = \ AD'+D'S'$
\end{itemize}
\vspace*{-0.3cm}

As $AB'$  is  parallel  to  $S'S''$  (both  in  frame  plane  and  both 
perpendicular to $l1$) we get: 
\vspace*{-0.3cm}
\begin{itemize}
\item[7)]
$ \frac{AD'}{D'S'} \  = \ \frac{AB'}{S'S''}$
\end{itemize}

\input{SZENE5}

\item
The shift of the whole  curve  from  the  projection  frame  in 
perpendicular direction (Fig.5.) has no effect on the shape  of  projection 
and hence may be omitted from consideration.
\end{enumerate}
\vspace*{-0.3cm}

     Remark: we have dropped index i on primed and  double  primed 
points and on S to increase the legibility of formulas. 

     Summarizing, we obtained 7 equations in unknowns:
\vspace*{-0.3cm}
\begin{quote}
 $c \ = \ AB, \ \alpha$ --- global for all frames

$\tau_{i}, \  \delta_{i}, \ AS', \ SS', \ D'S', \ S'S''$ --- local for a frame
\end{quote}
\vspace*{-0.3cm}

\noindent
(as $A'$, $B'$ and $D'$ are visible, so $AB' \ = \ c'_{i}$
 and $AD' \ = \ d'_{i}$ are known).

\noindent
We derive eliminating $AS$ by (1):
\vspace*{-0.3cm}
\begin{itemize}
\item[2$'$)]
$ AS' \ =  \ \cos \delta_{i}  * c * \mbox{tg}  \alpha$  and
\item[3$'$)]
$ SS' \ = \ \sin \delta_{i}  * c * \mbox{tg}  \alpha $
\end{itemize} 
\vspace*{-0.3cm}
Eliminating $AS'$ and $SS'$ by (2') and (3') we derive:
\vspace*{-0.3cm}
\begin{itemize}
\item[5$''$)]
$  S'S'' \ = \ \sin \tau_{i}  * \sin \delta_{i}  * c * \mbox{tg}  \alpha $ and
 \item[6$''$)]
$\cos \delta_{i}  * c * \mbox{tg}  \alpha \ = \ AD'+D'S'$
\end{itemize} 
\vspace*{-0.3cm}
Eliminating $S'S''$ and $D'S'$ by (5'') and (6'') we obtain:
\vspace*{-0.3cm}
\begin{itemize}
\item[7$'''$)]
$ \frac{AD'}{\cos \delta_{i} \ c \ \mbox{{\scriptsize tg}}  \alpha - AD'} \  
= \ \frac{AB'}
{\sin \tau_{i} \ \sin \delta_{i} \ \mbox{{\scriptsize tg}}   \alpha}$
\end{itemize} 
\vspace*{-0.3cm}
Substituting (4) into (7$'''$) we get:
\vspace*{-0.3cm}
\begin{itemize}
\item[8)]
$ d'_{i} \  c  \ \mbox{tg}  \alpha \ \sin \arccos (c'/c) \ \sin \delta 
 \ c'_{i} \  c \ \mbox{tg}  \alpha \  \cos \delta_{i}  - d'_{i} \  c'_{i}.$
\end{itemize} 
\vspace*{-0.3cm}
-- one equation with one local (frame dependent) unknown $\delta_{i}$.

\noindent
However, by analogy, we can derive the  second  equation  for  the 
same frame considering the opposite side of the  frame  plane  and 
the point $B$ and the tangential at $A$ instead of the point $A$ and the 
tangential at the point $B$. So we have the line $l2$ instead  of $l1$ 
crossing $B$, observable point $E'$ (and edge $BE' \ = \ e'_{i}$)
  instead  of  $D'$ 
(and $d'_{i}$). The rotation around $l2$ is the same as  around  $l1$  (i.e. 
$\delta_{i}$), but the rotation around $AB$ must be $\tau_{i} + \phi$, 
 $\phi$  being  the  angle 
between the plane containing $AB$ and tangential at $A$ and the  plane 
containing $AB$ and the tangential at $B$ (fixed for all  frames).  So 
we obtain:
\vspace*{-0.3cm}
\begin{itemize}
\item[9)]
$ e'_{i} \  c  \ \mbox{tg}\,\beta \ \sin \ \arccos (c'/c) \ \sin 
(\delta_{i} +\phi) \ =
\ c'_{i} \  c \ \mbox{tg}\,\beta \ \cos (\delta_{i} +\phi) \ 
 - \ e'_{i} c'_{i}$ .
\end{itemize} 
\vspace*{-0.3cm}
Let us introduce auxiliary (frame) terms,  containing  only  frame 
knowns and global unknowns:
\vspace*{-0.3cm}
\begin{quote}
$ q_{i1} \ = \ c'_{i} \  c \ \mbox{tg} \, \alpha ,  \ 
 p_{i1}\  = \ d'_{i}\  c \ \mbox{tg} \,\alpha  \sin \arccos (c'_{i}/c)$

$ q_{i2} \ = \ c'_{i} \  c \ \mbox{tg} \, \beta ,  \ 
 p_{i2}\  = \ e'_{i}\  c \ \mbox{tg} \, \alpha  \sin \arccos (c'_{i}/c)$
\end{quote}
\vspace*{-0.3cm}
So we have the equation system:
\vspace*{-0.3cm}
\begin{itemize}
\item[10)]
$ p_{i1} \sin \delta_{i} \ = \ q_{i1}  \cos \delta_{i} \  - \  d'_{i} c_{i}'.$

\item[11)]
$ p_{i2} \sin (\delta_{i} +\phi) \ = \ q_{i2} \cos (\delta_{i} +\phi) -  
e'_{i} c'_{i}$.
\end{itemize} 
\vspace*{-0.3cm}
Let us transform (10):\\
10$'$) $d'_{i} c'_{i}/ \sqrt{p_{i1}^{2}  +q_{i1}^{2}} \  = \ 
\cos \delta_{i} \  q_{i1} / \sqrt{p_{i1}^{2}  +q_{i1}^{2}} -
\sin \delta_{i} \ p_{i1}  / \sqrt{p_{i1}^{2}  +q_{i1}^{2}}$

\noindent
If we introduce a new auxiliary variable $ \omega_{i1} $   (with global 
unknowns only)
$$ \omega_{i1} \  =  \ \mbox{arc tg }(p_{i1}  /q_{i1} )$$
then we have:
\vspace*{-0.3cm}
\begin{itemize}
\item[10$''$)]
 $d'_{i} c'_{i}/ \sqrt{p_{i1}^{2}  +q_{i1}^{2}} \  = 
\ \cos (\delta_{i} +\omega_{i1}) $ and by analogy:
\item[11$''$)]
$e'_{i} c'_{i}/ \sqrt{ p_{i2}^{2}  +q_{i2}^{2}} \  = 
\ \cos (\delta_{i} + \phi + \omega_{i2}  )$
\end{itemize} 
\vspace*{-0.3cm}
and hence:
\vspace*{-0.3cm}
\begin{itemize}
\item[10$'''$)]
$ \arccos (d'_{i} c'_{i}/ \sqrt{p_{i1}^{2}  +q_{i1}^{2}} ) \ = 
\  \delta_{i} +\omega_{i1}$   and 
\item[11$'''$)]
$\arccos (e'_{i} c'_{i}/ \sqrt{ p_{i2}^{2}  +q_{i2}^{2}}) \ = 
\ \delta_{i} + \phi + \omega_{i2} $
\end{itemize} 
\vspace*{-0.3cm}
And thus we come to our final formula:

\vspace*{0.3cm}
\noindent
\fbox{ 12) $ \arccos (e'_{i} c'_{i}/ \sqrt{ p_{i2}^{2}  +q_{i2}^{2}}) 
- \arccos (d'_{i} c'_{i}/ \sqrt{p_{i1}^{2}  +q_{i1}^{2}} ) \  =
 \ \phi + \omega_{i2} - \omega_{i1} $ }
\vspace*{0.3cm}

\noindent
-- one equation for each frame in unknowns: $c$, $\alpha$, $\beta$ and 
$\phi$, 
which does not contain any frame dependent unknown. 

\noindent
For  determining  all  these  four  unknowns  characterizing   the 
reconstructed curve we need at least four frames.

\noindent
Degenerated cases (parallelism of lines) are  treated  easily  and 
will not be considered here.

     The formula (12) is, regrettably, not a practical one, though 
the equation system is solvable. Therefore the result is  more  of 
theoretical importance than  of  practical  one.  However,  it  is 
possible  to  transform 
this formula into a (high degree) polynomial in $c$, tg $\alpha$, 
tg $\beta$  and 
tg $\phi$, which can be a basis of a linear equation system constructed 
from a superfluous number of additionally observed  frames,  where 
the  solution  is  based  on  conjecture  of  linear   coefficient 
independence formulated in  \cite{K:10} and successfully  applied  therein 
to free motion under orthogonal projection  with  three  traceable 
points and 3+1 frames. \\ 
      Let us briefly outline the transformation of (12) into a lopynomial in 
the above-mentioned variables. After ''tangentializing'' and squaring twice 
we obtain  a quasi-polynomial of the form: \\  
 
\vspace*{-0.3cm}
\begin{itemize}
\item[12')]

$ s_{2}^{2} +s_{1}^{2} +{y}^{2}  + y  ^{2}  s_{2} ^{2}  s_{1} ^{2} - 
2s_{1}s_{2}-2ys_{2}-2ys_{2} ^{2} s_{1}- 
2ys_{1}-2ys_{2}s_{1} ^{2} -2y ^{2} s_{2}s_{1}-8ys_{2}s_{1} = 0$\\  

 \end{itemize}
\vspace*{-0.3cm}

with $y$ standing for $$y={tg}^2 (\phi + \omega_{i2} - \omega_{i1} )$$
and $s_{1}$, $s_{2}$ being proper polynomials: \\ 
$$ 
s_{1}= \frac
{ ({d'_{i}}^{2} +{c'_{i}}^{2} ){c}^{2}  tg 
\alpha -{d'_{i}} ^{2} {c'_{i}}^{2} ({tg}^{2}  \alpha-1) 
}  
{
 {d'_{i}}^{2} {c'_{i}}^{2}   
} $$  
 
$$ 
s_{2}= \frac
{ ({e'_{i}}^{2} +{c'_{i}}^{2} ){c}^{2}  tg 
\beta -{e'_{i}} ^{2} {c'_{i}}^{2} ({tg}^{2}  \beta -1) 
}  
{
 {e'_{i}}^{2} {c'_{i}}^{2}   
} $$

So the only non-polynomial factor is $y$. However:\\

$$
y=  \frac  
{ 
  (\frac { {d'_{i}} ^ {2}} 
        { {c'_{i}} ^ {2}}  
  - \frac  { {d'_{i}} ^ {2}} {{c} ^ {2} }
  )  
  ( 
   \frac { {e'_{i}} ^ {2}} { {c'_{i}} ^ {2}  }
    - \frac { {e'_{i}} ^ {2}} {{c} ^{2}     }  
  )
  ( {tg}  ^{2} \phi+1)   
  { 
    (1+ 
    \frac {{d_{i}} {e_{i}} } {{c'_{i}}  ^ {2}  }
    -\frac {{d_{i}} {e_{i}} } {{c} ^ {2}   }
    + tg \phi ( 
      \frac {{e_{i}} }{{c_{i}}  }
      -\frac {{e_{i}} }{{c_{i}} }
     ) \sqrt{1- 
     \frac {{c'_{i}} ^ {2}}{{c} ^ {2}  }
            } 
    )   
 }  
 ^  {2} 
} 
{  
{
({(1+{d_{i}} {e_{i}} /{c'_{i}}  ^{2}-{d_{i}} {e_{i}} /{c} ^ {2} )} ^ {2} -  
tg2 \phi ({e_{i}} /{c_{i}} -{d_{i}} /{c_{i}} ) ^{2} (1-{c'_{i}} ^ {2} /{c} ^ 
{2} ) )  
}  ^  {2}  
} -1  
$$ 

\vspace*{-0.3cm}

So we obtain an equation of the form: \\ 
$$polynomial1=polynomial2 * \sqrt(1-{c'_{i}} ^ {2} /{c} ^ {2} )$$
which is easily squared to obtain a proper polynomial in the above-mentioned 
variables. \\ To solve a system of equations being polynomials of high degree 
when superfluous observations from real world are  available we proceed the 
following way: we transform the equations in the following form:\\ 
$$ 0= \sum { expression-in-observables *  
product-of-variables-and-their-natural-powers} $$ 

We insist on each $product-of-variables-and-their-natural-powers$ be 
different in each summand. For each $product-...$ we introduce a new 
variable $a_{k}$ (something like the procedure when seeking a model for 
polynomial regression by means of linear regression method). In this way we 
obtain    a linear equation system which we solve using Gaussian method (if 
the number  of equations is equal to the number of new variables $a_{i}$ or 
by the least squares methods if the number of equations (that is, observed 
frames) is higher. \\ Solving such an equation system results in obtaining
another one with equations  of the form: $product of variables = constant$, 
which after application of logarithm results in a new linear equation system, 
this time in variables of primary interest.\\.
\\ Why should this method ( conjecture  of  linear   
coefficient independence) work ? Of course, degenerate cases are 
possible. It works however the very same way the linear regression does: we 
usually observe much less variables than there are degrees of freedom in the 
real world. 

\section{Remark on Reconstruction of Non-Traceable Points}

     The formulas of previous section allow to  identify  the  the 
relative (length c) as  well  as  the  absolute  position  of  the 
feature  points  $A.B$  as  well  as  the  (absolute  and  relative) 
direction of tangentials at A and B in space for each frame. 

     To recover the  shape  of  the  whole  smooth  curve,  it  is 
necessary to recover non-feature points also. It will be  possible 
only if points $A,B$ and tangentials are not co-planar. Then  each 
point in space is defined by means of parameters $(p1,p2,p3)$ as:
\vspace*{-0.3cm}
\begin{itemize}
\item[13)]
$ A+ \ p_{1} \ * \ AB \ + \ p_{2} \ * \ AC \  + \ p_{3} \ * \ AB \ \times  \ 
AC $\end{itemize} 
\vspace*{-0.3cm}
(the point $C$ be such that $AC$ is the tangential at $A$ and $AC$  is  of 
unit length, $x$ -- cross product indicator), and each straight  line 
as:
\vspace*{-0.3cm}
\begin{itemize}
\item[14)]
 $ f(u) \  = \   
p_{1} * AB \ + \ p_{2}  * AC \  + \ p_{3}  * AB \times  AC 
+ u* ( q_{1} * AB \ + \ q_{2}  * AC \  + \ q_{3}  * AB \times  AC)   $
\end{itemize} 
\vspace*{-0.3cm}
(O be the coordinate system origin)
It is obvious that in case of rigidly connected points $A,B,C$ every 
point and every straight line rigidly  connected  with  them  will 
retain the $p1,p2,p3(,q1,q2,q3)$  parameter  set  while  the  motion 
continues. 

\input{SZENE6}

     So let us  select  a  point  $X_{o}$   in  frame  $0$  lying  in  the 
projection  plane  on  the  projected  curve (Fig.6.),  this  point   being 
projection of a point $X$ of the curve. We  will  succeed  with  the 
reconstruction task for the point $X$ if we find  out  the  distance 
$XX_{o}$  for the frame $0.\, x$ be the name of the straight line connecting 
$X$ with $X_{o}$  (vertical to the projection plane). Let  us  obtain  the 
parameters $p1,p2,p3$ of this line and assume,  that  this  line  is 
fixed to the curve while moving. Let us draw now the projection of 
the line within the frame 1 obeying  the  function  $f(u)$  for  the 
frame $1$ position of $A,B,C$. Then this projected line will cross the 
projected curve at some points one of them being the  point  $X_{1}$-- 
the projection of our point $X$ of the curve. (On ambiguity  we  can 
recall the continuity of the curve). As we know  the  equation  of 
the straight line $XX_{1}$  in frame $1$ as well as that of  the  straight 
line $x$ we can easily recover the distance $XX_{1}$ , and  later  $XX_{0}$   
of the first frame. Proceeding in  this  way  we  recover  the  whole 
curve.(Ambiguities are resolved by continuity requirement). 

\section{Flat Curves in Prospective Projection - A Remark}

     This work profited from analysis  of  Lee's   \cite{L:12}  method  of 
reconstructing correspondence between two  orthogonal  projections 
of a flat curve in 3D. The basic idea there was  that  having  two 
traceable (end)points of the curve we have in fact three of  them: 
the third  being  the  crossing  point  of  tangentials  at  curve 
endpoints (as the curve is assumed  ``flat''  that  is  planar,  the 
tangentials -- unless parallel -- in fact have a common point). Then 
Lee simply exploited the Tales theorem in a straight forward way. 

\input{SZENE7}

     We would like to point out here that there is also a  similar 
simple method for reconstruction  of  correspondence  between  two 
PROSPECTIVE projections of a flat curve in 3D, but three traceable 
points of the curve are required then. Though no better  bound  is 
achieved for the number of traceable points required than that  in 
 \cite{K:11}, however the computational effort is drastically reduced:
Let the three traceable points be called  $A,B,C$  
(Fig.7.).  Clearly  usually 
the tangentials at $A$ and $B$ share a point, say $D$. Let us call $E$ the 
common point of straight lines $AB$ and $DC$. Let $A'$, $B'$, $C'$ $D'$, $E$  
be projections of $A,B,C,D,E$ respectively in the first frame, and  $A''$, 
$B'',\  C'',\  D'', \ E''$ be  respective  projections  in  the  second  
frame (Projections of $A,\ B$ and $C$ are visible, and projections of $D$ and 
 $E$ are easily obtainable by drawing).  Now  let  us  consider  $X'$,  a 
projection of the non-traceable point $X$ of the curve in the  first 
frame (let us select $X'$ freely on the curve  projection  image  of 
the first frame.). We want to find $X''$ being the projection of $X$ in 
the second frame. Let us call $Y$ the common point of lines  $AB$  and 
$DX$ --- its projection $Y'$ can be  obtained  by  drawing  as  crossing 
point of $A'B'$ and $D'X'$. Let us look for $Y''$ - the projection  of  $Y$ 
in the second frame. The well known elementary geometry theorem on 
prospective projection double quotient states that:
\vspace*{-0.3cm}
\begin{itemize}
\item[15)]
$\frac{AE}{AY} : \frac{BE}{BY} \  = \ \frac{A'E'}{A'Y'}: \frac{B'E'}{B'Y'} $  
\hspace{2cm}  and
\item[16)]
$ \frac{AE}{AY}: \frac{BE}{BY} \  = \ \frac{A''E''}{A''Y''}: 
\frac{B''E''}{B''Y''} $ \hspace{2cm} hence 
\item[17)]
$\frac{A'E'}{A'Y'}: \frac{B'E'}{B'Y'} \  = \ 
\frac{A''E''}{A''Y''}: \frac{B''E''}{B''Y''} $ 
\end{itemize} 
\vspace*{-0.3cm}
As the positions of the remaining points $A',  \ B',  \ E',  \ Y',
 \   A'',   \ B'', 
 \ E''$ are known, so based on (17)  $Y''$  is  easily  found  on  the  line 
$A''B''$. But $X''$ is the crossing point of the straight line  $D''Y''$  and 
the image of curve projection in the second frame, so easy to find 
Q.E.D.  (ambiguities  are  resolved  by  continuity  requirement). 
Degenerated cases (parallelism of lines) are  treated  easily  and 
will not be considered here.

\section{Conclusions}
     This paper makes two basic contributions to solution  of  the 
problem of reconstruction of rigid smooth curves from multiframes:
\vspace*{-0.3cm}
\begin{itemize}
\item[1.]
 decreases to 2 the theoretical lower bound  on  the  number  of 
   traceable points required to reconstruct the shape  of  a  true 
   3-D curve from multiframes  under  orthogonal  projection  with 
   totally unpredictable motion assumed  (the  previous bound  was 
   either 3 points  or  2  points  with  geometrical  or  physical 
   restriction on freedom of motion)
\vspace*{-0.3cm}
\item[2.]
 introduces a new algorithm  (based  on  double  quotient)  for 
   reconstruction of flat curves in 3  D  from  multiframes  under 
   prospective projection  using  3  traceable  points,  which  is 
   drastically simpler than that given in  \cite{K:11}. 
\end{itemize}
\vspace*{-0.3cm}
At this point the basic statement holding for  all  reconstruction 
algorithms based on multiframes should  be  repeated:  unless  the 
motion is a degenerate one (e.g. no motion at all, or no  rotation 
at all, or rotation around an  axis  perpendicular  to  the  frame 
plane etc.).
     If we compare the table given in Section 2 with  the  results 
of sections 3/4, we see easily that there is some ranking  on  the 
complexity of recovering algorithms depending on  the  amount  and 
type of information available. E.g. from  \cite{K:9} we know that  with  3 
traceable points and three frames available we obtain an  equation 
system with 3 mixed-quadratic equations in three  variables.  From 
 \cite{K:10} we know that adding one frame more leads us to   an  equation 
system with 3 linear equations in three  variables.  We  can  also 
observe that three point mean a special case of two points and two 
lines. From the complexity of equation (12) and the  fact  that  4 
frames are required at least, however, we see that availability of 
two lines is a much weaker information that that stemming  from  a 
third point. 
     Further research is  necessary  to  simplify  eventually  the 
solution given in (12). Also we hope that exploiting some insights 
from  consideration  of  flat  curves  in  3D  under   prospective 
projection also the bound of 4 traceable points necessary  by  now 
for true 3 D curves may be broken in future research work.

\end{document}